\documentclass[conference]{IEEEtran}
\IEEEoverridecommandlockouts
\usepackage[hidelinks]{hyperref}
\usepackage{cite}
\usepackage{amsmath,amssymb,amsfonts}
\usepackage{algorithmic}
\usepackage{graphicx}
\usepackage{textcomp}
\usepackage{xcolor}
\usepackage{cleveref}
\usepackage{tabularx}
\usepackage{bm}
\usepackage{booktabs} 
\usepackage{multirow}

\def\BibTeX{{\rm B\kern-.05em{\sc i\kern-.025em b}\kern-.08em
    T\kern-.1667em\lower.7ex\hbox{E}\kern-.125emX}}
\begin{document}

\title{MUTE-SLAM: Real-Time Neural SLAM with Multiple Tri-Plane Hash Representations}

\author{\IEEEauthorblockN{1\textsuperscript{st} Yifan Yan}
\IEEEauthorblockA{\parbox{2in}{\centering\textit{School of Automation Science\\ and Electrical Engineering}}\\
\textit{Beihang University}\\
Beijing, China \\
SY2203531@buaa.edu.cn}
\and
\IEEEauthorblockN{2\textsuperscript{nd} Ruomin He}
\IEEEauthorblockA{\parbox{2in}{\centering\textit{School of Automation Science\\ and Electrical Engineering}}\\
\textit{Beihang University}\\
Beijing, China \\
SY2203505@buaa.edu.cn}
\and
\IEEEauthorblockN{3\textsuperscript{rd} Zhenghua Liu}
\IEEEauthorblockA{\parbox{2in}{\centering\textit{School of Automation Science\\ and Electrical Engineering}}\\
\textit{Beihang University}\\
Beijing, China \\
lzh@buaa.edu.cn}
}
\maketitle

\begin{abstract}
We introduce MUTE-SLAM, a real-time neural RGB-D SLAM system employing multiple tri-plane hash-encodings for efficient scene representation. MUTE-SLAM effectively tracks camera positions and incrementally builds a scalable multi-map representation for both small and large indoor environments. As previous methods often require pre-defined scene boundaries, MUTE-SLAM dynamically allocates sub-maps for newly observed local regions, enabling constraint-free mapping without prior scene information. Unlike traditional grid-based methods, we use three orthogonal axis-aligned planes for hash-encoding scene properties, significantly reducing hash collisions and the number of trainable parameters. This hybrid approach not only ensures real-time performance but also enhances the fidelity of surface reconstruction. Furthermore, our optimization strategy concurrently optimizes all sub-maps intersecting with the current camera frustum, ensuring global consistency. Extensive testing on both real-world and synthetic datasets has shown that MUTE-SLAM delivers state-of-the-art surface reconstruction quality and competitive tracking performance across diverse indoor settings. The code is available at \url{https://github.com/lumennYan/MUTE_SLAM}.
\end{abstract}


\section{Introduction}
\label{sec:intro}
Dense Simultaneous Localization and Mapping (SLAM) has been a fundamental challenge in 3D computer vision for decades, playing a crucial role in applications like robotics, virtual/augmented reality, and autonomous driving. A robust dense SLAM system equipped with RGB-D sensors needs to track camera poses effectively while reconstructing the environment into a high-fidelity map.

Traditional dense SLAM methods \cite{engel2014lsd,newcombe2011dtam,newcombe2011kinectfusion,niessner2013real,whelan2012kintinuous,vespa2018efficient,zeng2013octree,dai2017bundlefusion,schops2019bad} generate accurate localization results and detailed 3D point positions, but they fall short in rendering novel views or filling unobserved regions. Learning-based systems \cite{huang2021di,bloesch2018codeslam,teed2021droid} have shown promise in large-scale scenes and global 3D map production, yet their reconstruction performance is limited and requires retraining for different scenarios.

With the advent of Neural Radiance Field (NeRF) \cite{mildenhall2021nerf}, efforts have been made to integrate it into SLAM systems due to its capability to render novel views \cite{barron2021mip,barron2022mip,barron2023zip,wang2023f2} and reconstruct 3D surfaces \cite{azinovic2022neural,or2022stylesdf,wang2021neus,li2023neuralangelo}. NeRF-based SLAM methods iMAP \cite{sucar2021imap} and NICE-SLAM \cite{zhu2022nice}, have demonstrated their applicability across various scenes and ability to predict the appearance of unobserved areas, although their computational demands hinder real-time application. Recent works \cite{wang2023co,jiang2023h2} utilize hash-encoded voxel grids \cite{muller2022instant} to accelerate convergence and enhance detail fidelity. Despite hash collisions can be implicitly mitigated by the original design of Instant-NGP \cite{muller2022instant}, the reconstructed mesh still suffers from aliasing. As proved in \cite{zhuang2023anti,johari2023eslam,ma2023otavatar,li2023efficient,ma2023otavatar}, projecting spatial points into a tri-plane reprensentation can efficiently reduce scene parameters while preserving geometric and color information. Therefore following \cite{li2023efficient}, we leverage a tri-plane hash-encoding method to store scene features and minimize hash collisions. Moreover, current Nerf-based SLAM systems \cite{sucar2021imap,zhu2022nice,johari2023eslam,wang2023co,Zhang_2023_ICCV} require pre-set scene boundaries, limiting their application in unknown environments. Some \cite{jiang2023h2,yang2022vox} address this with octree-based voxel grids, but they still necessitate an initially defined loose boundary and struggle to reconstruct beyond these limits. Although neural point cloud-based method \cite{sandstrom2023point} does not have such a concern, the point cloud-based representation requires large time and memory consumption, making it impractical for real-time usage. Our proposed MUTE-SLAM overcomes this by introducing a multi-map-based scene representation. Through dynamically allocating new sub-maps upon detecting new areas, MUTE-SLAM can be deployed in environments of any size, given reasonable RGB-D sensor observations, while maintaining reasonable runtime and memory overhead.

In summary, our contributions include:
\begin{itemize}[]
\item A multi-map-based scene representation facilitating reconstruction scalable to diverse indoor scenarios.
\item A tri-plane hash-encoding method for sub-maps which enables real-time tracking and anti-aliasing dense mapping with high-fidelity details.
\item A optimization strategy that jointly optimizes all sub-maps observed currently, ensuring global consistency.
\item Extensive experimental validation on various datasets, demonstrating our system's scalability and effectiveness in both tracking and mapping.
\end{itemize}

\begin{figure}[tb]
  \centering
  \includegraphics[width=0.5\textwidth]{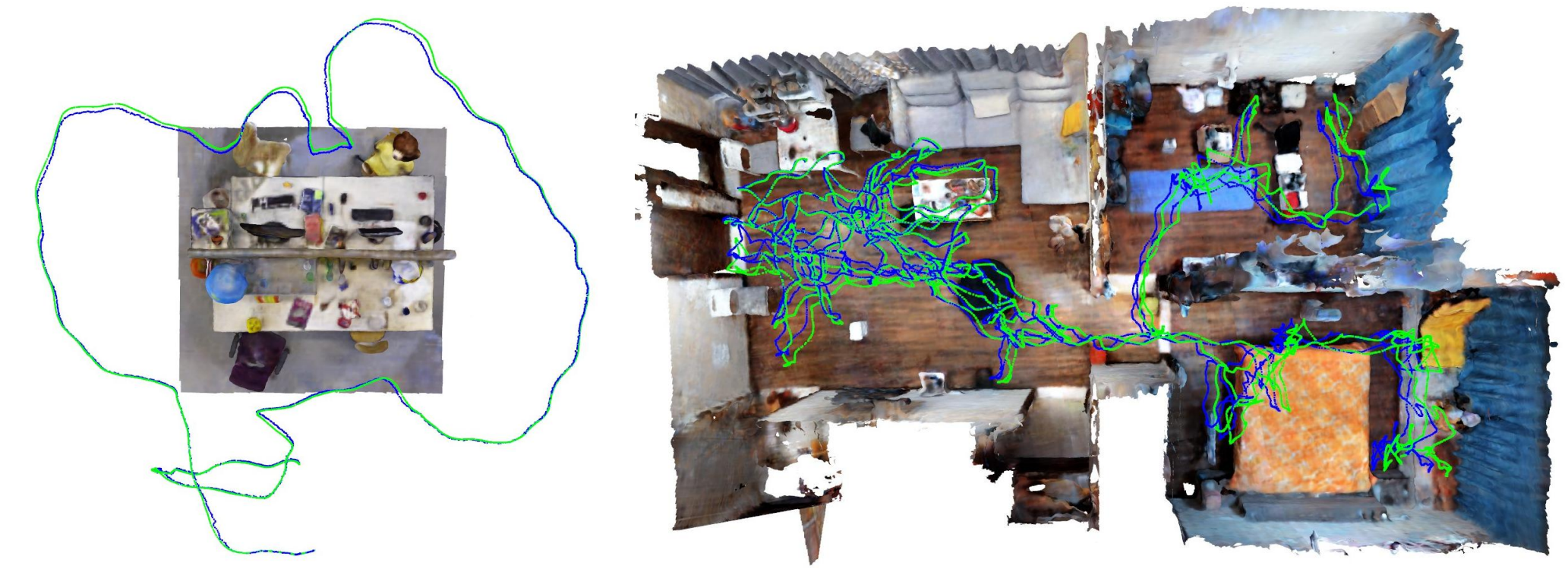}
  \caption{Our MUTE-SLAM system demonstrates rapid and accurate tracking and mapping across indoor environments of varying scales without pre-defined boundaries. We depict the trajectories and meshes of both a small and a large scenario: estimated trajectories are marked in blue, while ground truths are in green. The left image is an around-a-desk scene from the TUM-RGBD dataset \cite{sturm2012benchmark}, while the image on the right is 
 a multiple-room scene from Apartment dataset provided by NICE-SLAM \cite{zhu2022nice}.
  }
  \label{fig:first}
\end{figure}

\section{Related Works}
\subsection{Dense Visual SLAM}
DTAM \cite{newcombe2011dtam} is the first dense SLAM system to employ direct methods, utilizing information from all pixels for tracking by comparing newly inputted RGB frames with a reconstructed dense model. KinectFusion \cite{newcombe2011kinectfusion}, leveraging a RGB-D sensor, uses a volumetric Truncated Signed Distance Function (TSDF) to fuse scene geometry and tracks camera positions via Iterative Closest Point (ICP). Subsequent works have focused on improving scalability through new data structures \cite{zeng2013octree,niessner2013real}, enhancing global consistency with Bundle Adjustment (BA) \cite{schops2019bad,dai2017bundlefusion}, and increasing efficiency \cite{whelan2012kintinuous}. Recent learning-based methods \cite{bloesch2018codeslam,teed2021droid,huang2021di} demonstrate superior accuracy and robustness compared to traditional approaches on single scenes, but they struggle with generalization across varied scenes.

\subsection{Neural Implicit SLAM}
iMAP \cite{sucar2021imap} is the first SLAM system to incorporate NeRF \cite{mildenhall2021nerf}, it models the environment into a single Multilayer Perceptron (MLP) and jointly optimizes the map and camera poses. NICE-SLAM \cite{zhu2022nice} substitutes the scene representation in iMAP with hierarchical voxel girds to resolve the forgetting issue, achieving enhanced tracking and mapping in large indoor environments. However, both approaches face limitations in handling unknown environments due to the requirement for a prior scene boundary. Vox-Fusion \cite{yang2022vox} attempts to address this by introducing octree-based voxel grids as the map representation but is still limited to the initially defined spatial scope. 

Recent advancements in NeRF-based SLAM have improved tracking, mapping, and running speed. ESLAM \cite{johari2023eslam} stores features on multi-scale axis-aligned planes and employs rendering based on TSDF. Co-SLAM \cite{wang2023co} combines coordinate and hash-encodings for input points to achieve both smooth meshes and fast convergence, and introduces a real-time global bundle adjustment mechanism using a ray list sampled from past keyframes. GO-SLAM \cite{Zhang_2023_ICCV} and H2-mapping \cite{jiang2023h2} combine traditional SLAM modules with neural mapping, achieving high tracking accuracy. Despite these improvements, scalability remains a challenge. Our work addresses this by introducing a multi-map solution with an accompanying optimization strategy, which requires no pre-set boundaries. Besides, we represent each sub-map with a tri-plane hash-encoding which allows for fast convergence and detailed surface reconstruction. 

\begin{figure*}[tb]
  \centering
  \includegraphics[trim={0cm 0cm 0cm 0cm},clip,height=9cm]{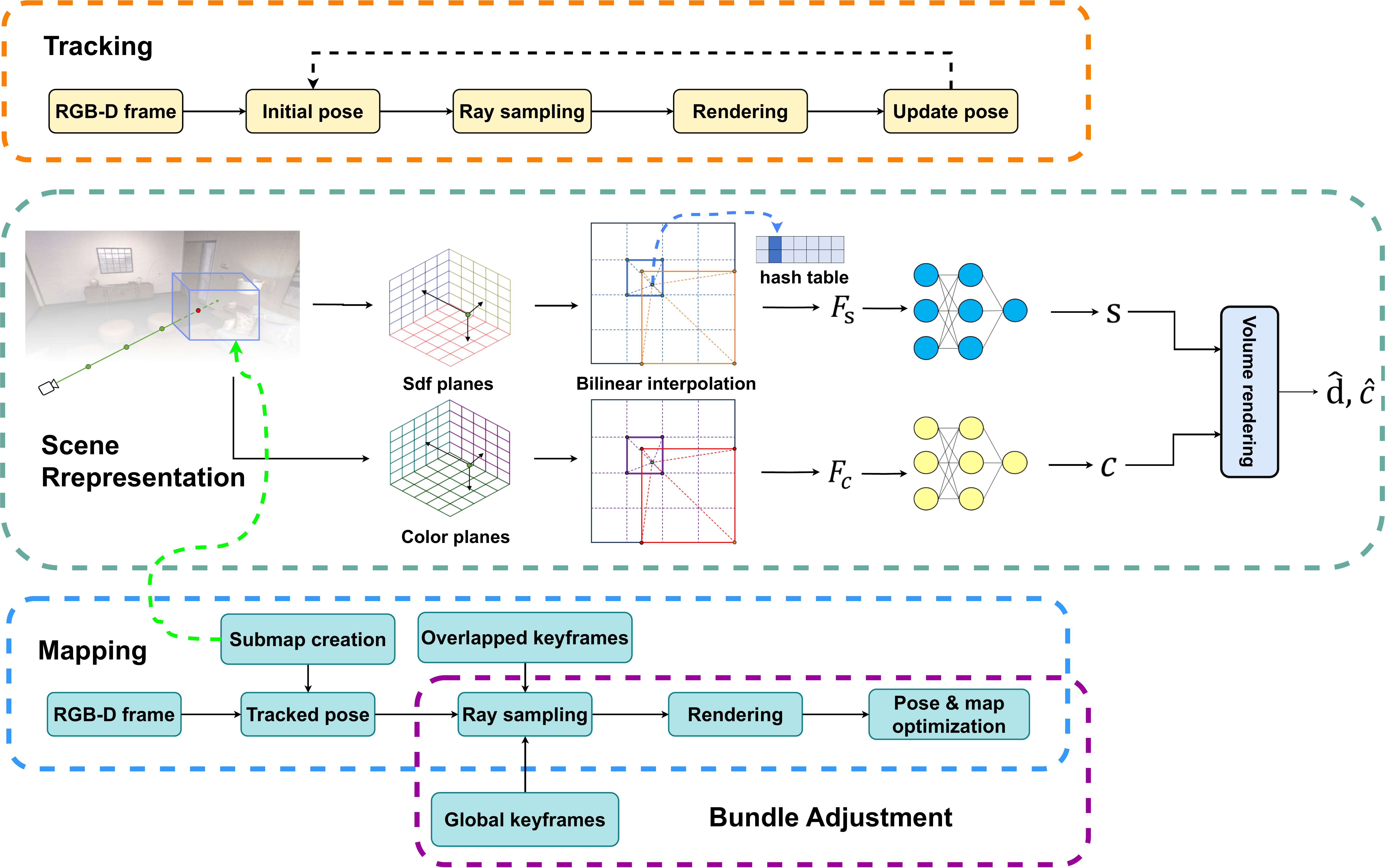}
  \caption{The overview of MUTE-SLAM.Our method consists of three parts. 1)Scene representation: the whole scene is represented by several sub-maps created on the fly. Each sub-map is formulated by double tri-plane hash-encoders, one for TSDF and the other for color encoding. 2)Tracking: this module optimizes the pose for each frame through differentiable rendering. 3)Mapping: the mapping module dynamically allocates new sub-maps with a tracked pose. It conducts a joint optimization of both scene and pose parameters, utilizing the current frame along with co-visible keyframes. 4)Bundle Adjustment: by sampling keyframes globally, this module further refines all trainable parameters and ensures global consistency.}
  \label{fig:system_overview}
\end{figure*}

\section{Method}
The overview of MUTE-SLAM is illustrated in \cref{fig:system_overview}. Given an input RGB-D stream $\{I_i, D_i\}_{i=1}^M$, the system tracks their 6-DOF camera poses $\{T_i:R_i|t_i\}_{i=1}^M$ and conducts mapping to optimize a multi-map implicit scene representation $\{E_k\}_{k=1}^S$. 
When the camera captures a new region, a corresponding sub-map is created to represent it, as detailed in \cref{sec:Multi-map}.
Each local map $E_k$ encodes a point coordinate $\bm{p}$ within its domin with three orthogonal TSDF feature planes and three color planes, as described in \cref{sec:Tri-plane}.
These features from the sub-maps are decoded into TSDF and color using two separate MLPs, initiating the volume rendering process (\cref{sec:Rendering}). The rendered depth and color images are subsequently utilized to jointly optimize the sub-maps and camera poses. Additionally, periodic global bundle adjustments are implemented to ensure global consistency(\cref{sec:T&M}).
\subsection{Multi-map Scene Representation}
\label{sec:Multi-map}
As previous neural implicit SLAM methods \cite{sucar2021imap,zhu2022nice,yang2022vox,johari2023eslam,wang2023co} are restricted to functioning within pre-defined scene boundaries, they are unsuitable for navigating and mapping large, unknown indoor environments. Consequently, performing tracking and mapping incrementally with no prior environment information becomes a critical issue. MUTE-SLAM addresses this problem by adopting a multi-map scene representation approach. We encode the whole scene with several sub-maps $\{E_k\}_{k=1}^S$, each to express a local region, enabling the reconstruction of indoor scenes of arbitrary shapes and sizes.

After tracking an input RGB-D frame, points are randomly sampled from the depth image and projected into the world coordinate system with the estimated camera pose $\{T_i:R_i|t_i\}$. Points with invalid depths are filtered out, and outliers are removed to mitigate noise. If the proportion of points that fall outside all existing sub-maps exceeds a predetermined threshold $P$, a new local map $E_k$ is generated. The local map's size is determined by extending the cuboid vicinity of the current camera position and the points that are out of bounds over a length of $l$, which is a hyperparameter.  The redundancy of a sub-map's boundary would reduce the number of total sub-maps, thereby lowering memory consumption. Concurrently, the corresponding frame is added to the global keyframe database. For optimization, rays are sampled from the current frame and co-visible keyframes. Rays terminating outside all sub-maps are removed from the training process. Specifically, rather than optimizing a single sub-map at a time, we simultaneously optimize all observed sub-maps to ensure global consistency as a input frame's frustum may intersect with multiple sub-maps. Global bundle adjustments are also employed periodically to further enhance global consistency.

\subsection{Tri-plane Hash Encoding}
\label{sec:Tri-plane}
Lately, hash-encoding \cite{muller2022instant} has gained much attention in the NeRF community \cite{barron2023zip,jiang2023h2,li2023neuralangelo,wang2023co} due to its fast convergence and strong environmental representation capabilities. Despite that, hash collisions are inevitable and could lead to artifacts in the reconstructed scenes. The standard solution is to let the light MLP decoder handle hash collisions implicitly or use larger hash tables. But as the scene grows large and complicated, the MLP will reach its limit and a capacious hash table consumes substantial memory space. Meanwhile, tri-plane encoding approaches \cite{zhuang2023anti,johari2023eslam,ma2023otavatar,li2023efficient,ma2023otavatar} have demonstrated competence in surface and appearance reconstruction with low memory consumption. Combining the advantages of both worlds, we represent each sub-map by tri-plane hash-encoding.

In MUTE-SLAM, a local map $E_k$ is encoded by three orthogonal planes for TSDF $\{E_{xy}^s, E_{xz}^s, E_{yz}^s\}$ and another three for color$\{E_{xy}^c, E_{xz}^c, E_{yz}^c\}$, each plane denotes a 2D hash encoder as in \cite{muller2022instant}. All planes share the same resolution levels $L$, base resolution $N_{min}$, finest resolution $N_{max}$, per level feature dimension $\chi$ and hash table size $H$. The finest resolution $N_{max}$ and hash table size $H$ are determined by the local map volume $V [m^3]$:
\begin{align}
    N_{max}&= \lfloor 50\cdot V^{\frac{1}{3}} \rfloor \label{eq:max_resolution} \\
    H&=N_{max}^2 \label{eq:hashtable}
\end{align}
Other parameters are set as hyperparameters. When a point falls within a sub-map, it is orthogonally projected onto the corresponding planes.  The encoder then interpolates the point features bilinearly by querying the nearest four vertices from each level of the hash table, concatenating features across all levels to produce the final output. Consequently, the TSDF and color feature vectors $\{F_s, F_c\}$ are derived by summing up the outputs from the three planes:
\begin{align}
  F_s(\bm{p}) = E_{xy}^s(\bm{p}) + E_{xz}^s(\bm{p}) + E_{yz}^s(\bm{p})\\
  F_c(\bm{p}) = E_{xy}^c(\bm{p}) + E_{xz}^c(\bm{p}) + E_{yz}^c(\bm{p})
  \label{eq:feature}
\end{align}
We employ two separate double-layer MLPs $\{f_s, f_c\}$ to decode the TSDF and RGB values respectively:
\begin{align}
  s(\bm{p}) = f_s(F_s(\bm{p}))\\
  \bm{c(\bm{p})} = f_c(F_c(\bm{p}))
  \label{eq:feature}
\end{align}
The TSDF and color values are then utilized in the volume rendering module.
The plane-based representation, growing quadratically with scene size, results in fewer hash table queries and hence fewer collisions compared to grid representations, given equal hash table sizes. Furthermore, since a point's feature vector is a composite of inputs from three distinct encoders, the probability of encountering conflicts across all encoders is significantly reduced. In situations where collisions do occur in one encoder, the impact is mitigated by the inputs from the other encoders, thus lessening the overall adverse effect. \cref{ablations} demonstrates the effectiveness of our tri-plane hash-encoding.
\subsection{TSDF-based Volume Rendering}
\label{sec:Rendering}
MUTE-SLAM follows the TSDF-based rendering procedure in \cite{johari2023eslam}. For an input frame $i$, we sample randomly from pixels with valid depths, casting rays into the world coordinate system using the estimated pose $\{T_i:R_i|t_i\}$. We apply stratified sampling to acquire $N=N_g+N_d$ points along a ray, distributed between the $near$ and $far$ bounds. Initially, $N_g$ points are uniformly sampled across the entire sampling region. Then within a smaller truncated distance $[d-\delta, d+\delta]$ near the surface, extra $N_d$ points are sampled uniformly, where $d$ denotes the ground truth depth value. 

Each sampled point $\bm{p_n}=\bm{o}+d_n\bm{r}$ is represented by the ray's origin $\bm{o}$, direction $\bm{r}$ and depth $d_n$. For all points $\{\bm{p_n}\}^N_{n=1}$ along a ray, we predict the color $\bm{\hat{c}}$ and depth $\hat{d}$ with color and TSDF values $\{\bm{c(\bm{p_n})},s(\bm{p_n})\}$ retrieved from MLPs:
\begin{equation}
    \bm{\hat{c}} = \sum_{n=1}^{N}\omega_n\bm{c(\bm{p_n})},\ 
    \hat{d} = \sum_{n=1}^{N}\omega_n d_n
  \label{eq:render}
\end{equation}
For $\omega_n$, as in \cite{or2022stylesdf}, it is derived from the volume density $\sigma(\bm{p_n})$ and TSDF value $s(\bm{p_n})$:
\begin{gather}
    \sigma(\bm{p_n}) = \beta\cdot Sigmoid(-\beta\cdot s(\bm{p_n}))\\
    \omega_n = exp(-\sum_{k=1}^{n-1}\sigma(\bm{p_k}))\cdot[1-exp(-\sigma(\bm{p_n}))]
   \label{eq:omega}
\end{gather}
Here, $\beta$ is a learnable parameter which controls the sharpness of surfaces.

\subsection{Tracking and Mapping}
\label{sec:T&M}
\subsubsection{Loss Functions.}
We apply four loss functions to optimize the scene representation, MLPs and camera poses: RGB loss, depth loss, TSDF loss and free-space loss. Once a batch of rays $R$ are selected, the RGB and depth loss are obtained as $l_2$ errors between rendered and ground truth values:
\begin{gather}
    L_{rgb} = \frac{1}{|R|}\sum_{r \in R}(\bm{\hat{c}(r)}-\bm{c(r)})^2\\
    L_{depth} = \frac{1}{|R|}\sum_{r \in R}(\bm{\hat{d}(r)}-\bm{d(r)})^2
   \label{eq:render_loss}
\end{gather}
The free-space loss is applied to supervise the points far from the surfaces ($|d_p-d(r)|>\delta$) to have a truncated TSDF value of $1$:
\begin{equation}
    L_{fs} = \frac{1}{|R|}\sum_{r \in R}\frac{1}{|N_{fs}|}\sum_{\bm{p} \in N_{fs}}(s(\bm{p})-1)^2
   \label{eq:freespace_loss}
\end{equation}
For points near the surface ($|d_p-d(r)|<=\delta$), similar to \cite{johari2023eslam}, we further split them into two parts to obtain the TSDF loss:

\begin{gather}
    L_{mid} = \frac{1}{|R|}\sum_{r \in R}[\frac{\lambda_{mid}}{|N_{mid}|}\sum_{\bm{p} \in N_{mid}}(s(\bm{p})+d(r)-d_p)^2]\\
    L_{tail} = \frac{1}{|R|}\sum_{r \in R}[\frac{\lambda_{tail}}{|N_{tail}|}\sum_{\bm{p} \in N_{tail}}(s(\bm{p})+d(r)-d_p)^2]\\
    L_{tsdf} = L_{mid} + L_{tail}
   \label{eq:tsdf_loss}
\end{gather}

Where the middle points $\bm{p} \in N_{mid}$ with depths  reside in $[d(r)-0.4\delta,d(r)+0.4\delta]$ have a larger weight $\lambda_{mid}$, while the others have a smaller weight $\lambda_{tail}$.
The final loss is the weighted sum of the objective functions above:
\begin{equation}
    L_{all} = \lambda_{rgb}\cdot L_{rgb}+\lambda_{depth}\cdot L_{depth}+\lambda_{fs}\cdot L_{fs}+L_{tsdf}
   \label{eq:loss_all}
\end{equation}
\subsubsection{Tracking.}
We track the camera-to-world transformation matrix $T_i\in SE(3)$ for every input frame $i$. When receiving a input frame $i$, its initial pose is obtained using constant speed assumption:
\begin{equation}
    T_i = T_{i-1}T_{i-2}^{-1}T_{i-1}
   \label{eq:pose_intial}
\end{equation}
Then, the pose $T_i$ is transformed into a seven-dimensional vector for optimization, which is formed by concatenating the rotation quaternion $q_i$ and the translation vector $t_i$. We sample uniformly $N_{cam}$ pixels from frame $i$ and optimize the pose iteratively using all loss functions, while keeping the scene parameters and MLPs fixed.
\subsubsection{Mapping.}
MUTE-SLAM performs mapping every $K$ frames and inserts the mapped frame as a keyframe into the global keyframe database. When the mapping thread starts, we first sample $N_{map}$ rays from the current frame and $M$ keyframes having co-visibility with current frame. Then we filter out rays whose $far$ bounds lay outside all sub-maps. Each point on the rays is encoded within the corresponding sub-map. Since we define loose boundaries for sub-maps, only the oldest map is used when points fall into the area where multiple sub-maps overlap. At last, we jointly optimize all observed sub-maps, MLPs and camera poses iteratively with the objective functions. Specifically, we use the ground truth pose at the first input frame and only optimize scene parameters and MLPs for initialization. 
\subsubsection{Bundle Adjustment.}
Once the keyframe database has accumulated a sufficient number of frames, global bundle adjustment is initiated for every twenty frames of input. From the keyframe database, $G$ frames are globally sampled, with all trainable parameters optimized in a manner akin to the mapping thread. The global bundle adjustment module plays a crucial role in correcting drifting poses and bolstering global consistency.

\section{Experiments.}
\subsection{Experimental Setup.}

\subsubsection{Baselines}
We choose state-of-art NeRF-based dense SLAM approaches ESLAM \cite{johari2023eslam}, Co-SLAM \cite{wang2023co} and Point-SLAM \cite{sandstrom2023point} as our main baselines for both surface reconstruction and camera tracking. To better evaluate our proposed MUTE-SLAM on pose estimation, we also compare with previous methods NICE-SLAM \cite{zhu2022nice} and Vox-Fusion \cite{yang2022vox}. We run these methods using the default settings provided in their open-source code. We do not compare with\cite{jiang2023h2} and \cite{Zhang_2023_ICCV} as they both use traditional SLAM modules for camera poses estimation.

\subsubsection{Datasets}
We evaluate MUTE-SLAM on various 3D benchmarks of indoor scenarios. For quantitative evaluation of the reconstruction quality, we use 8 synthetic scenes from Replica \cite{straub2019replica}. To validate effectiveness on pose tracking, we conduct experiments on 6 real-world scenes from ScanNet \cite{dai2017scannet} and 3 real-world scenes from TUM-RGBD \cite{sturm2012benchmark} dataset. We also demonstrate the scalability of MUTE-SLAM on large-scale Apartment dataset provided by NICE-SLAM \cite{zhu2022nice}.

\subsubsection{Metrics}
For surface reconstruction, We adopt four evaluation metrics: $Depth \ L1\ [cm]$, $Accuracy\ [cm]$, $Completion\ [cm]$ and $Completion\ ratio \ [<5cm\%]$. Additionally, to underscore the ability of our method to produce detailed geometry compared to ESLAM \cite{johari2023eslam}, we also incorporate the $Completion\ ratio \ [<1cm\%]$ metric. Following \cite{azinovic2022neural,wang2022go,johari2023eslam}, before evaluation, we remove faces that are not inside any camera frustum or are occluded in all RGB-D frames from the reconstructed mesh. For the evaluation of camera tracking, we employ $ATE\ RMSE\ [cm]$ \cite{sturm2012benchmark}.

\subsubsection{Implementation Details}

We run all experiments on a desktop PC with a 3.70GHz Intel i9-10900K CPU and an NVIDIA RTX 3080 GPU. For local map creation, the threshold $P$ is set to 0.2 for Replica \cite{straub2019replica} and 0.25 for the other datasets, while the expanding size $l$ is 1 m for Replica \cite{straub2019replica}, 1.5 m for ScanNet \cite{dai2017scannet}, 2.5m for Apartment \cite{zhu2022nice} and 3 m for TUM-RGBD \cite{sturm2012benchmark}. Each hash encoder has the same base resolution $N_{min}=16$, resolution levels $L=16$, per level feature dimension $\chi=2$, resulting in 32 dimensions of input for MLPs. The MLPs both have two hidden layers of 32 channels. For rendering, we set the near-surface truncated distance $\delta$ to 6 cm and regular sampling number $N_g$ to 32. Particularly, we sample $N_d = 8$ near-surface points for Replica \cite{straub2019replica} and TUM-RGBD \cite{sturm2012benchmark}, and $N_d=12$ points for ScanNet \cite{dai2017scannet} and Apartment \cite{zhu2022nice}. Please refer to the supplementary materials for further details of our implementation. 

\subsection{Evaluation of Mapping and Tracking}
\subsubsection{Evaluation on Replica \cite{straub2019replica}}
\begin{figure*}[tb]
  \centering
  \includegraphics[trim={0cm 0cm 0cm 0cm},clip,height=8cm]{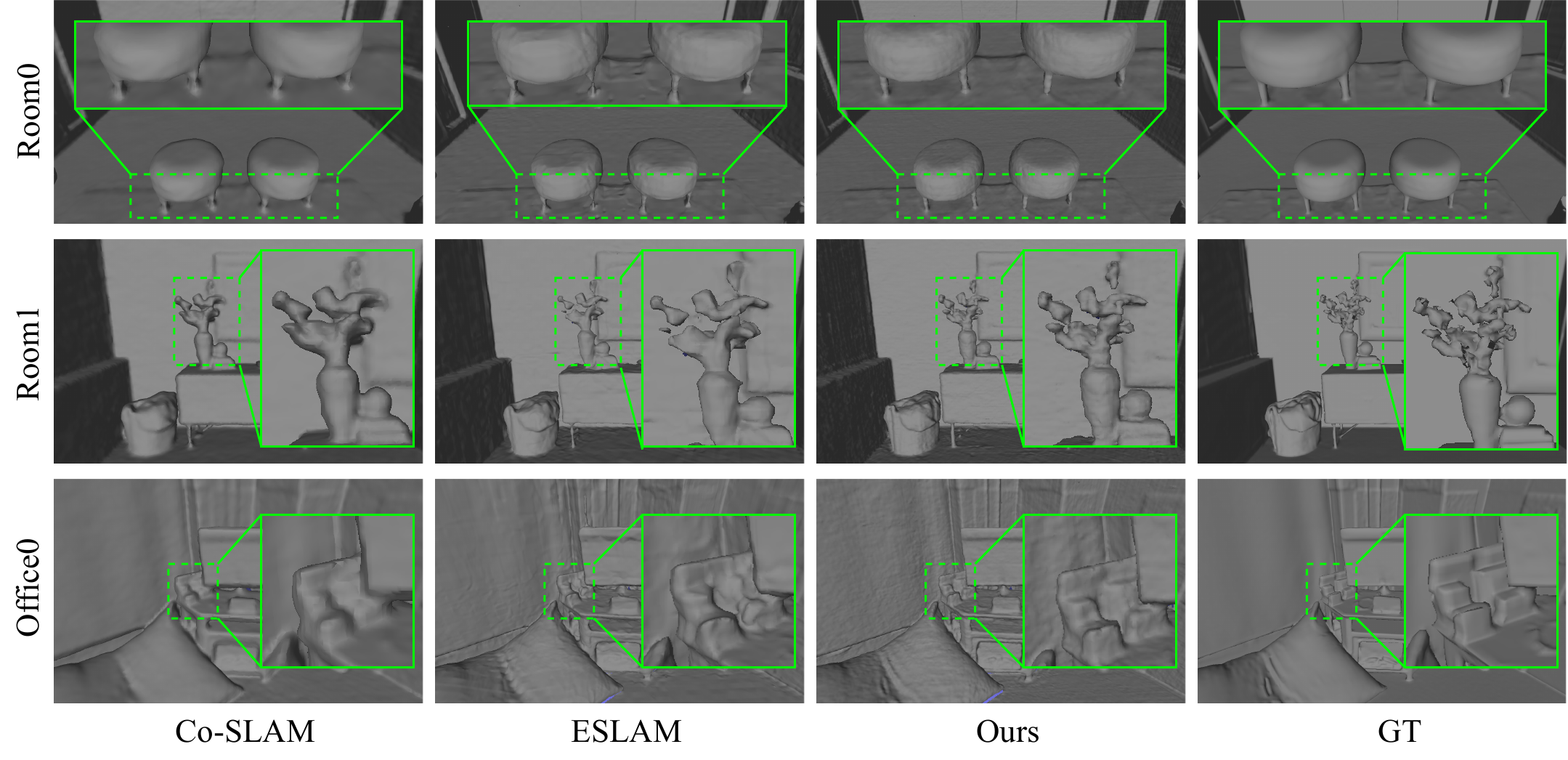}
  \caption{Qualitative reconstruction results on Replica.}
  \label{fig:replica}
\end{figure*}

We compare the reconstruction performance on Replica \cite{straub2019replica} only with Co-SLAM \cite{wang2023co}, Point-SLAM \cite{sandstrom2023point} and ESLAM \cite{johari2023eslam} as they significantly outperform previous methods \cite{sucar2021imap,zhu2022nice,yang2022vox}. However, the origin setting of Point-SLAM takes hours on Replica, which is unfair for other baselines as they only require a few minutes. Thus we modified the number of rays to 2000 and iterations to 10 for Point-SLAM, and denote the modified version as $\text{Point}^{\ast}$. Note that $\text{Point}^{\ast}$ \cite{sandstrom2023poin} still takes two times longer than other methods. For quantitative analysis, we run each method five times and report the average results. As shown in \cref{tab:replica}, our approach outperforms Co-SLAM \cite{wang2023co} and $\text{Point}^{\ast}$ \cite{sandstrom2023point} on all scenes and shows competitive performance with ESLAM \cite{johari2023eslam}. Due to the use of a joint coordinate and parametric encoding, Co-SLAM \cite{wang2023co} tends to produce over-smoothed surfaces, which leads to the amplification of reconstruction error. Point-SLAM \cite{sandstrom2023point} can reconstruct fine scene contents as in the origin setting, but is impractical in real-time use. Although ESLAM \cite{johari2023eslam} achieves high overall accuracy, it falls short in preserving surface details. To further highlight our method's superiority in capturing scene details, we compare the $Completion\ ratio \ [<1cm\%]$ with ESLAM \cite{johari2023eslam} in \cref{tab:replica_detail}. Qualitative results in \cref{fig:replica} also demonstrate that MUTE-SLAM effectively reconstructs detailed environmental geometry with fewer artifacts. Because the modified version $\text{Point}^{\ast}$ \cite{sandstrom2023poin} performs poorly, we do not show its qualitative results.

\begin{table}[tb]
  \caption{Quantitative results of reconstruction on Replica \cite{straub2019replica} dataset.}
  \label{tab:replica}
  \centering
  \scriptsize
  \renewcommand{\arraystretch}{1.2}
  \begin{tabular}{@{}c|cccc@{}}
    \toprule
    Method & Depth L1 (cm) $\downarrow$ & Acc. (cm) $\downarrow$ & Comp. (cm) $\downarrow$ & Cp. Ratio (\%) $\uparrow$ \\
    \hline
    ESLAM & 1.19 & 0.94 & 0.96 & 99.31\\
    Co-SLAM & 3.22 & 1.18 & 1.12 & 98.49\\
    $\text{Point}^{\ast}$ & 8.39 & 3.72 & 2.10 & 94.09\\
    Ours & \textbf{1.18} & \textbf{0.91} & \textbf{0.95} & \textbf{99.34}\\
   \bottomrule
  \end{tabular}
\end{table}

\begin{table}[tb]
  \caption{Comparison of $Completion\ ratio \ [<1cm\%]$ with ESLAM\cite{johari2023eslam}.}
  \label{tab:replica_detail}
  \centering
  \scriptsize
  \renewcommand{\arraystretch}{1.2}
  \begin{tabularx}{\linewidth}{@{}p{0.7cm}|XXXXXXXXX@{}}
    \toprule
     Method & Room0 & Room1 & Room2 & Office0 & Office1 & Office2 & Office3 & Office4 & Avg.\\
    \hline
    ESLAM & 55.21 & 72.77 & 65.94 & 76.30 & 86.17 & 62.90 & 49.05 & 54.55 & 65.36\\
    Ours & \textbf{56.29} & \textbf{74.55} & \textbf{66.62} & \textbf{76.93} & \textbf{88.61} & \textbf{64.28} & \textbf{49.51} & \textbf{55.67} & \textbf{66.56}\\
  \bottomrule
  \end{tabularx}
\end{table}

\subsubsection{Evaluation on ScanNet\cite{dai2017scannet}}
\begin{table}[tb]
  \caption{Quantitative results of ATE RMSE (cm) on ScanNet \cite{dai2017scannet}. }
  \label{tab:scannet}
  \centering
  \begin{tabular}{@{}c|ccccccc@{}}
    \toprule
    SceneID & 0000 & 0059 & 0106 & 0169 & 0181 & 0207 & Avg.\\
    \hline
    NICE-SLAM & 8.61 & 12.24 & 8.04 & 10.27 & 13.01	 & 5.55 & 9.62\\
    Vox-Fusion & 8.42 & 9.20 & \textbf{7.42} & 6.60 & 12.13 & \textbf{5.51} & 8.21\\
     ESLAM & 7.37 & 8.31 & 7.59 & 6.45 & \textbf{9.29} & 5.65 & \textbf{7.44}\\
     Co-SLAM & 7.77 & 12.52 & 9.39 & 6.34 & 12.35 & 7.65 & 9.34\\
     Point-SLAM & 10.24 & \textbf{7.81} & 8.65 & 22.16 & 14.77 & 9.54 & 12.19\\
     Ours& \textbf{7.08} & 9.07 & 8.27 & \textbf{6.18} & 10.21 & 7.19 & 8.00\\
  \bottomrule
  \end{tabular}
\end{table}

\begin{figure*}[tb]
  \centering
  \includegraphics[width=0.85\textwidth]{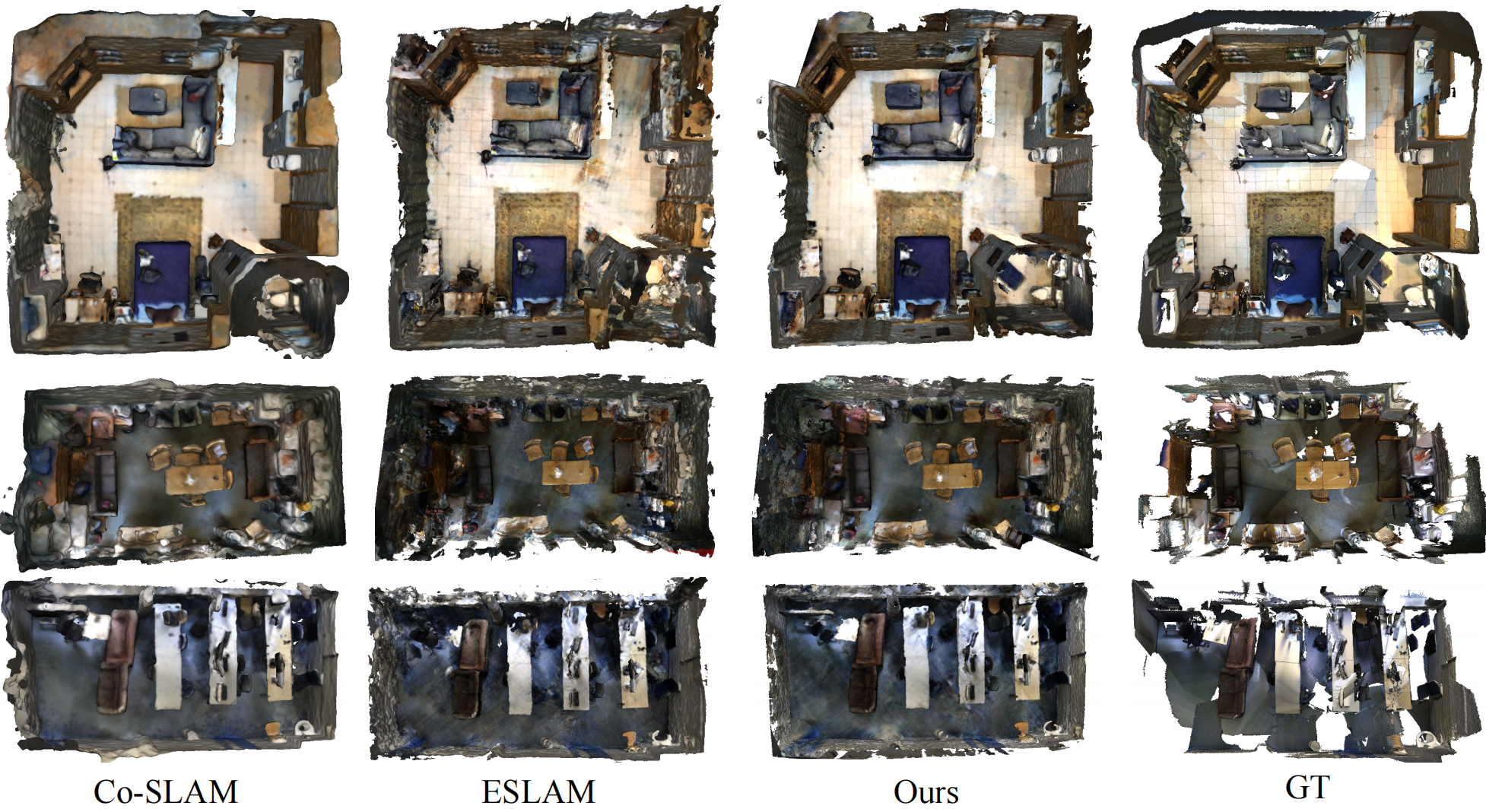}
  \caption{Qualitative reconstruction results on ScanNet \cite{dai2017scannet}. Our reconstructed mesh achieves better completion and fewer artifacts compared to ESLAM \cite{johari2023eslam}. Additionally, our method produces sharper and more detailed geometry than Co-SLAM \cite{wang2023co}.}
  \label{fig:scannet}
\end{figure*}

We assessed camera tracking accuracy on six real-world scenes from the ScanNet dataset \cite{dai2017scannet} and report the average ATE RMSE \cite{sturm2012benchmark} across five runs for each scene and method in \cref{tab:scannet}. Our approach, MUTE-SLAM, exhibits competitive results in these tests. Notably, even without pre-defined scene boundaries, MUTE-SLAM consistently outperforms Co-SLAM \cite{wang2023co} in all tested scenes. While ESLAM \cite{johari2023eslam} achieves the best overall performance, it does so with twice the average processing time as ours (\cref{tab:performance}). Our approach not only surpasses ESLAM in several scenes but also secures the second-best overall result.Due to the incomplete nature of ScanNet \cite{dai2017scannet} meshes, we present only qualitative reconstruction results in \cref{fig:scannet}. These results highlight MUTE-SLAM's ability to capture finer details and achieve a high level of completeness in reconstructions.

\subsubsection{Evaluation on TUM RGB-D \cite{sturm2012benchmark}}
\begin{table}[tb]
  \caption{Quantitative results of ATE RMSE (cm) on TUM RGB-D\cite{sturm2012benchmark}.}
  \label{tab:tum}
  \centering
  \begin{tabular}{c|ccc}
    \toprule
     Method & fr1/desk(cm) & fr2/xyz(cm) & fr3/office(cm)\\
    \hline
     NICE-SLAM  & 2.7 & 1.8 & 3.0 \\
     ESLAM  & \textbf{2.5} & 
     N/A & \textbf{2.8}\\
     Co-SLAM & 2.9 & 1.8 & 2.9\\
     Ours & 2.7 & \textbf{1.3} & \textbf{2.8}\\
     \hline
     BAD-SLAM & 1.7 & 1.1 & 1.7\\
     Kintinuous & 3.7 & 2.9 & 3.0\\
     ORB-SLAM2  & \textbf{1.6} & \textbf{0.4} & \textbf{1.0}\\
  \bottomrule
  \end{tabular}
\end{table}
To further evaluate the tracking accuracy of MUTE-SLAM, we conducted experiments on real-world scenes from the TUM RGB-D dataset \cite{sturm2012benchmark}, with results averaged over five runs.  Instance of failure is denoted as ‘N/A’. Noted that ESLAM \cite{johari2023eslam} runs unsuccessfully on the 'fr2/xyz' scene and has been consequently excluded from these comparative results. As shown in \cref{tab:tum}, our quantitative analysis reveals that MUTE-SLAM not only outperforms NICE-SLAM \cite{zhu2022nice} and Co-SLAM \cite{wang2023co} but also demonstrates competitive performance and superior robustness compared to ESLAM \cite{johari2023eslam}, which takes hours to run on this dataset while ours only takes a few minutes. We also compare with some traditional SLAM methods \cite{schops2019bad, whelan2012kintinuous, mur2017orb} on this dataset. While NeRF-based SLAM methods still lag behind them, MUTE-SLAM narrows the gap. 

\subsubsection{Evaluation on Apartment \cite{zhu2022nice}}
To demonstrate the effectiveness of our method in large scale indoor scenarios, we evaluate the tracking and surface reconstruction performance on Apartment dataset provided by NICE-SLAM \cite{zhu2022nice}.  The failed instance is denoted as N/A. As illustrated in \cref{tab:apartment}, our method yields reasonable tracking performance. It should be emphasized that our method runs the fastest on this dataset, as discussed in \cref{performance_analysis}.
\begin{table}[tb]
  \caption{Quantitative results of ATE RMSE (cm) on Apartment \cite{zhu2022nice} dataset.}
  \label{tab:apartment}
  \centering
  \begin{tabular}{c|ccccc}
    \toprule
     Method & NICE.  & Vox.  & ESLAM  & Co. & Ours\\
     \hline
     ATE RMSE (cm) $\downarrow$ & 5.66 & 12.84 & N/A & 6.73 & 6.97\\
  \bottomrule
  \end{tabular}
\end{table}

\subsection{Performance Analysis}
\label{performance_analysis}
\begin{table}[tb]
  \caption{Run-time and memory comparison on Replica \cite{straub2019replica}, ScanNet \cite{dai2017scannet}, and Apartment \cite{zhu2022nice} scenes.}
  \label{tab:performance}
  \centering
  \begin{tabular}{c|c|cc}
    \toprule
      & Method & Speed FPT(s) & \# Param.\\
      \hline
     \multirow{3}{*} {Replica} & ESLAM & 0.18 & 6.85M\\
     & Co-SLAM  & \textbf{0.12} & \textbf{0.26M}\\
     & Ours & 0.21 & 6.28M\\
     \hline
     \multirow{3}{*} {ScanNet} & ESLAM  & 0.56 & 17.8M\\
     & Co-SLAM  & \textbf{0.19} & \textbf{1.59M}\\
     & Ours & 0.28 &10.73M\\
     \hline
    \multirow{3}{*} {Apartment} & ESLAM  & 2.40 & 22.1M\\
     & Co-SLAM  & 0.23 & \textbf{1.59M}\\
     & Ours & \textbf{0.22} & 12.38M\\
  \bottomrule
  \end{tabular}
\end{table}
We conducted a comparative analysis of the speed and memory consumption between our proposed MUTE-SLAM and the state-of-the-art methods ESLAM \cite{johari2023eslam} and Co-SLAM \cite{wang2023co}. Evaluations were performed on diverse scales of scenes: the small-scale 'room0' from Replica \cite{straub2019replica}, the mid-scale '0000' from ScanNet \cite{dai2017scannet}, and the large-scale Apartment scene from NICE-SLAM \cite{zhu2022nice}. Our metrics included average frame processing time (FPT) and the model's parameter count. As shown in \cref{tab:performance}, MUTE-SLAM not only operates faster in large scale scenes but also requires less memory compared to ESLAM \cite{johari2023eslam}. Notably, in large-scale scenarios like Apartment \cite{zhu2022nice}, MUTE-SLAM achieves even speed advantages over Co-SLAM \cite{wang2023co}. Moreover, the FPT and memory usage of MUTE-SLAM remain relatively stable across scene sizes, a benefit attributable to our scene representation design. Although Co-SLAM's \cite{wang2023co} coordinate hash encoding reduces runtime and memory usage, its smoothing effect hinders detailed scene reconstruction. Therefore, coordinate hash encoding is preferable when fine-grained reconstructions are unnecessary, whereas our tri-plane hash offers a better balance of overhead and performance for detailed needs.
\subsection{Ablations}
\label{ablations}
\begin{table}[tb]
  \caption{Quantitative results of ablation study. }
  \label{tab:ablations}
  \centering
  \scriptsize
  \begin{tabular}{c|cccc|c}
    \toprule
     \multirow{2}{*}{Method} & \multicolumn{4}{c|}{Replica} & ScanNet\\
     & Acc.$\downarrow$& Comp.$\downarrow$ & Cp. Ratio$\uparrow$ & Depth L1$\downarrow$ & ATE$\downarrow$ \\
     \hline
     w/o multi-map & 1.04 & 1.00 & 99.34 & 1.20 & 9.64\\
     w/o tri-plane & \textbf{0.96} & \textbf{0.98} & 99.30 & 1.16 & 9.48\\
     w/o global BA & 1.06 & 1.08 & 99.17 & 1.27 & 12.06\\
     ours full & 1.03 & 1.01 & \textbf{99.74} & \textbf{0.93} & \textbf{8.00}\\
  \bottomrule
  \end{tabular}
\end{table}

\begin{figure*}[tb]
  \centering
  \includegraphics[width=0.9\textwidth]{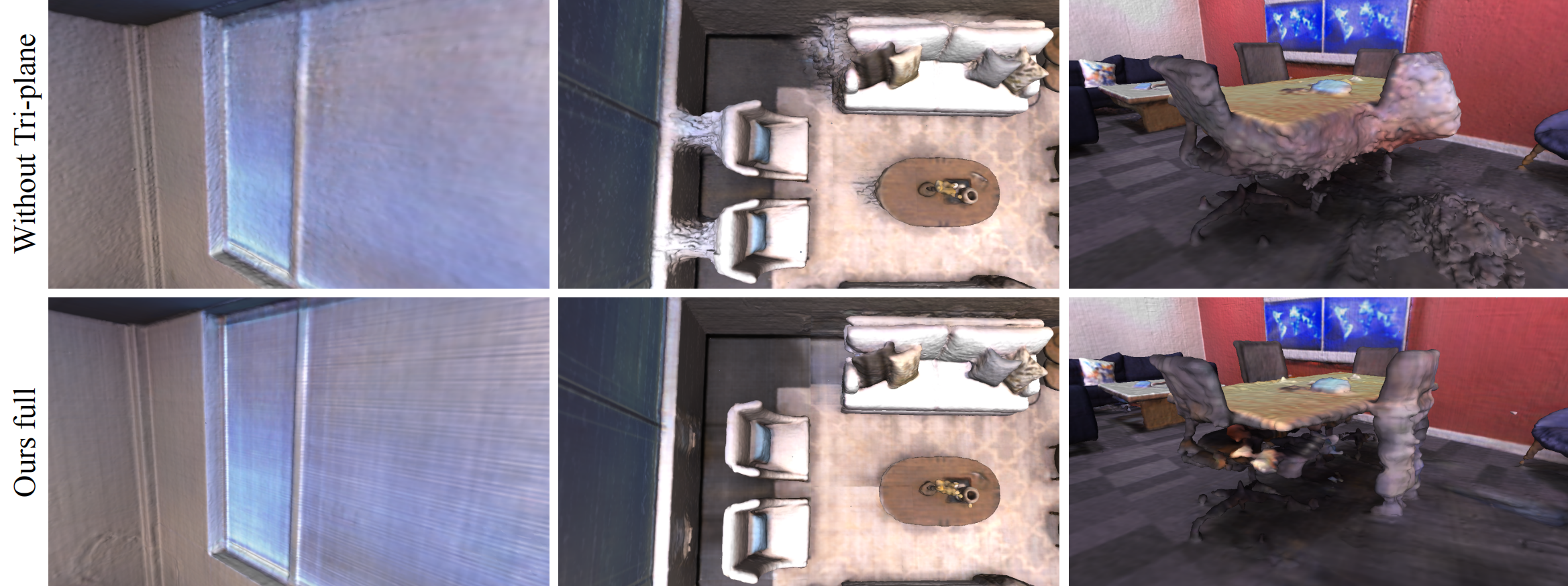}
  \caption{Qualitative comparison of our method employing tri-plane hash-encoding versus without it, using reconstructed meshes from Replica \cite{straub2019replica} scenes.The left-most images illustrate how hash collisions can result in rough surfaces and low-quality textures in flat areas like walls and windows. Our tri-plane approach significantly mitigates these issues, achieving better results even with smaller hash tables. The other two images further show that our design leaves fewer artifacts in unobserved regions.
  }
  \label{fig:wo_tri}
\end{figure*}

\subsubsection{Multi-map representation}
We conducted ablation experiments on the Replica \cite{straub2019replica} and ScanNet \cite{dai2017scannet} datasets to evaluate the impact of various components of our design. The quantitative results of this study are detailed in \cref{tab:ablations}. Representing the scene with one map, our findings indicate that the multi-map representation improves tracking performance. This enhancement in tracking accuracy, in turn, leads to higher quality in the reconstruction process. This improvement can be attributed to our strategy for allocating submaps. By extending the boundaries of each submap over a defined length, the corresponding hash tables are able to attain larger sizes, which contributes to better overall system performance. To further prove that ill-set scene boundaries harm performances of mapping and tracking, we conducted a experiment on the Replica \cite{straub2019replica} 'room0' sequence for all baselines in the need of pre-set scene boundaries as in \cref{tab:bound}. To simulate the scenario where the camera exits the pre-set boundary, we modify the boundary to a cube with a side length of 10m for all baselines (except for Vox-Fusion due to its implementation) and report the 'origin / ill-set' results. Note that NICE-SLAM fails on this setting. The results indicate significant degradation for all baselines, meanwhile our multi-map representation will not encounter such a problem.
\begin{table}[tb]
  \caption{Impact of ill-set scene boundary. }
  \label{tab:bound}
  \centering
  \tiny
  \begin{tabular}{c|cccc|c}
    \toprule
     Method & Depth L1 $\downarrow$ & Acc. $\downarrow$ & Comp. $\downarrow$ & Cp. Ratio $\uparrow$ & ATE $\downarrow$ \\
    \hline
     NICE-SLAM & N/A & N/A & N/A & N/A & N/A \\
     Vox-Fusion (64*0.2m) & 1.12 / 3.45 & 1.21 / 1.48 & 1.35 / 1.61 & 93.85 / 90.42 & 1.2 / 6.7 \\
     ESLAM & 0.95 / 49.74 & 1.04 / 1.11 & 1.03 / 33.97 & 99.71 / 72.04 & 0.67 / 9.3 \\
     Co-SLAM & 1.47 / 52.09 & 1.04 / 1.24 & 1.06 / 31.47 & 99.45 / 72.74 & 0.63 / 0.64 \\
  \bottomrule
  \end{tabular}
\end{table}
\subsubsection{Tri-plane hash-encoding}
To assess the effectiveness of tri-plane hash-encoding, we conducted an experiment where we replaced the tri-plane in each hash-encoding with a grid, while simultaneously tripling the maximum hash table size $N_{max}$. This adjustment marginally increases the overall capacity of the hash tables compared to the tri-plane approach. \cref{tab:ablations} shows that tri-plane hash-encoding achieves superior tracking results, a higher completion ratio, and better-rendered depth images. Although grid hash-encoding excels in terms of accuracy and completion, it leads to artifacts in the reconstructed mesh due to hash collisions, which in turn affects tracking accuracy. As illustrated in \cref{fig:wo_tri}, our qualitative comparison demonstrates that our proposed tri-plane hash-encoding effectively reduces aliasing and preserves scene details more accurately.
\subsubsection{Global bundle adjustment}
As illustrated in \cref{tab:ablations}, the lack of global bundle adjustment leads to higher ATE errors and relatively low reconstruction performance. As global bundle adjustment corrects drifting poses and refines scene representation, it plays a critical role in ensuring robustness and global consistency in our method.
\section{Conclusion}
We presented MUTE-SLAM, a dense real-time neural RGB-D SLAM system utilizing multiple tri-plane hash-encodings as scene representation. We demonstrate that utilizing several sub-maps to express the scene ensures scalability, making our method applicable to various indoor scenarios. We also show that integrating tri-plane with hash-encoding diminishes hash collisions and trainable parameters, producing high-fidelity surface reconstruction and low memory usage. Moreover, we perform global bundle adjustment periodically to achieve accurate poses estimation and maintain global consistency.
\section{Limitations.} 
Our method relies on the valid observation of RGB-D sensors, thus is susceptible to illumination changes and inaccurate depth measurements. Additionally, our approach of randomly sampling from all historical keyframes for global bundle adjustment might result in insufficient optimization in less frequently observed regions, potentially compromising reconstruction quality in these areas.
\bibliographystyle{IEEEtran}
\bibliography{IEEEexample}
\end{document}